\documentclass[10pt,twocolumn,letterpaper]{article}

\usepackage{cvpr}              % To produce the CAMERA-READY version
\usepackage{graphicx}
\usepackage{amsmath}
\usepackage{amssymb}
\usepackage{booktabs}
\usepackage{balance}
\usepackage{epigraph}
\usepackage{paralist}
\usepackage{multirow}
\usepackage[accsupp]{axessibility}
\usepackage{mathptmx}
\usepackage[11pt]{moresize}
\usepackage[table,xcdraw]{xcolor}

\newcommand{\mcircle}[1]{\raisebox{1pt}{\textcircled{\raisebox{-.9pt} {#1}}}}
\newlength\savewidth\newcommand\shline{\noalign{\global\savewidth\arrayrulewidth
  \global\arrayrulewidth 1pt}\hline\noalign{\global\arrayrulewidth\savewidth}}
\newcommand{\tablestyle}[2]{\setlength{\tabcolsep}{#1}\renewcommand{\arraystretch}{#2}\centering\footnotesize}

\definecolor{citecolor}{HTML}{0071bc}
\usepackage[pagebackref=true,breaklinks=true,letterpaper=true,colorlinks,citecolor=citecolor,bookmarks=false]{hyperref}

\usepackage{xcolor}

\usepackage[capitalize]{cleveref}
\crefname{section}{Sec.}{Secs.}
\Crefname{section}{Section}{Sections}
\Crefname{table}{Table}{Tables}
\crefname{table}{Tab.}{Tabs.}

\begin{document}

\title{\textsc{GanOrCon:} Are Generative Models Useful for Few-shot Segmentation?}

\author{Oindrila Saha \quad \quad Zezhou Cheng \quad \quad  Subhransu Maji\\
University of Massachusetts Amherst\\
{\tt\small \{osaha, zezhoucheng, smaji\}@cs.umass.edu}
}

\twocolumn[{
\renewcommand\twocolumn[1][]{#1}
\maketitle
\begin{center}
\vspace{0mm}
    \centering
  	\captionsetup{type=figure, width=.95\linewidth}
	\hspace{-6mm}
	    \includegraphics[width=0.95\linewidth]{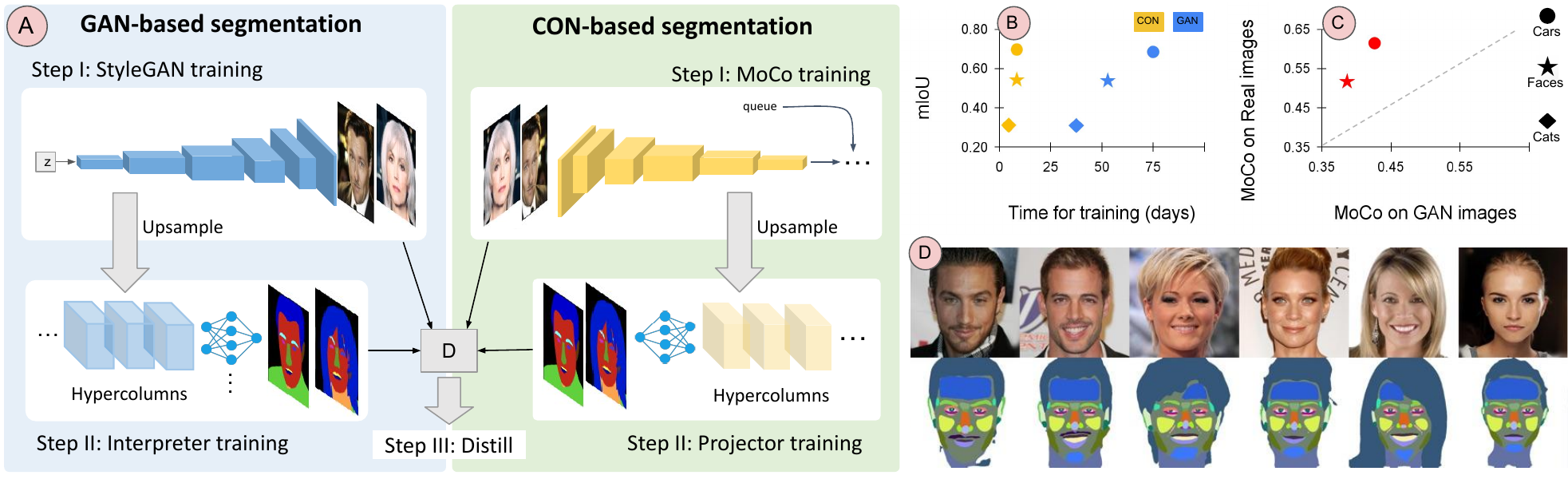} 
	\vspace{-3mm}
    \caption{\textbf{\textsc{GanOrCon} --- A comparison of GAN and contrastive learning (CON) approaches for few-shot part segmentation.} \mcircle{A} Both approaches consist of three steps. \textbf{Step~I:} Train a GAN decoder for generating images or a contrastive learning based image encoder. \textbf{Step~II}: Train a projector given a few labeled examples using hypercolumn representations from the decoder or encoder. \textbf{Step~III:} For efficient inference, distill the GAN sampled images and their labels to an off-the-shelf feed-forward segmentation model. This step is optional for CON as the model is feed-forward. \mcircle{B} CON representations achieve better performance while being nearly an order-of-magnitude faster to train than GAN representations on several datasets. \mcircle{C} CON representations are significantly worse when trained on GAN generated images. \mcircle{D} CON representations are effective at fine-grained part segmentation tasks --- the figure shows the output of our method trained with 16 labeled faces.}
    \label{fig:splash}
\vspace{1mm}
\end{center}
}]
\maketitle
%%%%%%%%% ABSTRACT
\begin{abstract}
\vspace{-2mm}
Advances in generative modeling based on GANs has motivated the community to find their use beyond image generation and editing tasks.
In particular, several recent works have shown that GAN representations can be re-purposed for discriminative tasks such as part segmentation, especially when training data is limited.
But how do these improvements stack-up against recent advances in self-supervised learning?
Motivated by this we present an alternative approach based on contrastive learning and compare their performance on standard few-shot part segmentation benchmarks.
Our experiments reveal that not only do the GAN-based approach offer no significant performance advantage, their multi-step training is complex, nearly an order-of-magnitude slower, and can introduce additional bias.
These experiments suggest that the inductive biases of generative models, such as their ability to disentangle shape and texture, are well captured by standard feed-forward networks trained using contrastive learning.
\end{abstract}

%%%%%%%%% BODY TEXT
\section{Introduction}
\label{sec:intro}
\vspace{-2mm}
\setlength\epigraphwidth{0.9\linewidth}
\setlength\epigraphrule{0pt}
\epigraph{``When solving a problem of interest, do not solve a more general problem as an intermediate step."}{---\emph{Vladimir Vapnik}}
\vspace{-4mm}

Generative modeling has long been used for unsupervised representation learning for recognition tasks.
But recently, several works~(e.g., \cite{zhang2021datasetgan,tritrong2021repurposing}) have shown that representations from off-the-shelf GANs (e.g.,~\cite{brock2018large,karras2019style}) can be re-purposed for part segmentation tasks, in part due to their ability to generate highly realistic images.
These models can be trained with a few labeled samples, often outperforming transfer learning baselines for a range of object categories.
Yet, generating images is arguably more challenging than learning representations for part labeling. 
This begs the question if the intermediate step of training a GAN is useful. What information, if any, is lost in the process of training a GAN, and if alternatives, in particular those based on unsupervised constrastive learning (e.g.,~\cite{he2019momentum,chen2020simple,chen2021exploring}), are just as effective.
Despite recent advances in generative and contrastive learning, a careful comparison of their effectiveness for few-shot part segmentation is lacking in the literature.
We thus begin by presenting a strong contrastive learning baseline for this task.
We then identify the key steps in these approaches and presents a series of experiments to evaluate their importance.

The two schemes are illustrated in Fig.~\ref{fig:splash} and described in \S\ref{sec:method}.
Generative modeling approaches~\cite{zhang2021datasetgan,tritrong2021repurposing} first train a generator $\mathbf{x}=f(\mathbf{z})$ on the unlabeled samples $\mathbf{x} \sim \{\mathbf{x}_i\}_{i=1}^n$ and $\mathbf{z} \sim N(0, \sigma^2\mathbb{I})$. Once trained a few generated samples are labeled by an annotator to train a part label predictor $h$.
This is used to generate a large labeled dataset $(\mathbf{x},\mathbf{y}) \sim \{(\mathbf{x}_i, \mathbf{y}_i)\}_{i=1}^m$ and ``distilled" to a feed-forward segmentation network $g$ using the dataset as a source of supervision.
Constrastive learning (CON) techniques train a network $f(\mathbf{x})$ to encode an image using an objective that maximizes the similarity of images under transformations (e.g.,~\cite{dosovitskiy2014discriminative,he2019momentum,chen2020simple}). Once trained a classifier $h$ is trained to predict part labels given labeled examples on top of $f$. The combined model $h \circ f$ can be optionally distilled to another network $g$.
While the details vary, we identify three factors that affect the overall performance on the task:
\begin{compactenum}
    \item Complexity -- how long does is it to train the model and how many steps are involved?
    \item Effectiveness -- how effective are the learned representations on the few-shot segmentation task. In particular how does the choice of the labeling network $h$ and distillation step affect the overall performance.
    \item Robustness -- How does the differences in distribution of GAN generated data from the real data affect labeling difficulty and performance.
\end{compactenum}

Our experiments suggest that GAN representations offer no significant advantage over CON representations in each of these dimensions.
GAN training is nearly an order-of-magnitude slower (Fig.~\ref{fig:splash} and \S\ref{sec:design}).
Across four datasets (Face-34, Face-8, Cat16, Car20) and two evaluation metrics (IoU and Weighted IoU), CON representations consistently outperform GAN representations, as well as strong baselines based on transfer learning and semi-supervised learning (Tab.~\ref{tab:usvsthem}).
GANs performance is especially poor when measured directly using latent code optimization to embed the test images, but distillation to a feed-forward segmentation network improves their generalization significantly. Yet, their performance remains below CON representations (Tab.~\ref{tab:detailedcomparison} \& \ref{tab:lco}).
CON representations see a smaller benefit from distillation unlike the GAN representations, suggesting that a significant portion of the generalization ability of GANs can be attributed to distillation to the final segmentation network.
CON representations also generalize better when trained on domain specific datasets than ImageNet (Tab.~\ref{tab:pretraining}).
Remarkably, we find that while the performance of the final segmentation network trained on real and synthetically generated data are comparable, but CON representations trained on GAN generated images perform significantly worse (Tab.~\ref{tab:generatedvsreal}).
This might indicate that GAN generated imagery contain artifacts that makes instance discrimination artificially easy (See \S\ref{sec:results} for details), and puts caution to their use for fine-gained classification tasks.

In summary our contributions are: 1) a strong contrastive learning approach for part segmentation. We evaluate the effectiveness learning frameworks (MoCo~\cite{he2019momentum} vs SimSiam~\cite{chen2020simple} in Tab.~\ref{tab:mocovssimsiam}) and architectures for predicting part segmentations on top of the learned representations. The approach outperforms recent StyleGAN-based approaches on standard few-shot part segmentation benchmarks. 2) An analysis of the GAN and CON representations in terms of the three factors listed above. We find that GAN-based approaches take longer to train, are less accurate, and suffer from labeling difficulty due to their inability to generate fine-grained parts.
These experiments suggest that the inductive biases of GANs, such as their ability to disentangle shape and texture, are well captured by standard feed-forward networks trained using contrastive learning. The source code and data associated with the paper are publicly available at \href{https://people.cs.umass.edu/~osaha/ganorcon}{https://people.cs.umass.edu/$\sim$osaha/ganorcon}.

\vspace{-5pt}
\section{Related Work}
\vspace{-3pt}
\textbf{Generative models} learn a probability distribution using unlabeled samples and extract representations from the learned distribution $p(\mathbf{x}|\theta)$.
For example, Jaakkola and Haussler~\cite{jaakkola1999exploiting} map data points $\mathbf{x}$ based on Fisher score $U_\mathbf{x} = \nabla_\theta \log p(\mathbf{x}|\theta)$, enabling the use of Hidden Markov Models to represent DNA sequences. Fisher vector representations~\cite{sanchez2013image,simonyan2013fisher} have been used to encode the distribution of local descriptors in images (e.g., SIFT~\cite{lowe1999object}, LBP~\cite{ojala1994performance}) using Gaussian mixture models~\cite{mclachlan1988mixture}.
RBMs~\cite{smolensky1986information}, VAEs~\cite{kingma2013auto}, GANs~\cite{goodfellow2014generative}, among others employ deep networks to learn distributions over complex, high-dimensional signals representing images, 3D shapes, speech, and text. Once trained, the internal representations of these networks can be used for discriminative tasks.

Besides their use in image generation and editing tasks, the ability to generate highly realistic data has motivated the community to explore GANs as a source of training data for discriminative tasks. 
Besnier et al.~\cite{besnier2020dataset} train image classifiers based on class-conditional samples from BigGAN~\cite{brock2018large} and identify heuristics to improve generalization.
Jahanian et al.~\cite{jahanian2021generative} use GANs to sample views of data to learn representations in a contrastive learning framework, while Tanaka and Aranha~\cite{tanaka2019data} use GANs for data augmentation.
While much of prior work has focused on image classification, two recent works, DatasetGAN~\cite{zhang2021datasetgan} and ReGAN~\cite{tritrong2021repurposing}, show that hypercolumn representations~\cite{hariharan2015hypercolumns} from StyleGANs~\cite{karras2019style} are effective for fine-grained image segmentation.
These representations generalize well when a small number of labeled examples are provided, and 
the model can be distilled to an off-the-shelf image segmentation model by training it on a large number of GAN-generated images and their labels.

Yet, generating data can be more challenging than learning the desired invariances directly. Self-supervised learning provides an alternative for learning these invariances from unlabeled data.
These techniques are based on training deep networks to solve a proxy task derived form the data alone.
While early works proposed tasks such as image colorization~\cite{Zhang2016}, orientation prediction~\cite{Gidaris2018}, jigsaw puzzle tasks~\cite{Noroozi2016}, and inpainting~\cite{Pathak2016}, \textbf{constrastive learning}~\cite{dosovitskiy2014discriminative, hjelm2018learning,oord2018representation,bachman2019learning,Wu2018a,Tian2019,he2019momentum,chen2020simple,chen2020improved,chen2020big} has emerged as the leading technique.
These techniques are based on applying synthetic geometric and photometric image transformations and maximizing the mutual information between the transformations of a single image based on variations of the InfoNCE loss~\cite{oord2018representation}.
Contrastive learning has been shown to be an effective alternative to ImageNet pre-training for various downstream tasks.
While the goal is often to learn a global image representation, several works have proposed variants for dense image prediction tasks such as part segmentation or landmark detection~\cite{cheng2021equivariant,wang2021dense, pinheiro2020unsupervised, thewlis2019unsupervised}.
The dense variants are based on extracting hypercolumns across layers of a network to obtain spatially varying representations, or modifying the contrastive learning objective to account for different locations within an image, e.g., by adding a within-image contrastive loss.

The goal of this work is to compare and contrast GAN-based and contrastive learning (CON) based approaches for few-shot segmentation based on their complexity, effectiveness and robustness and highlight some of the challenges in each. 
GANs act as a compact representation of the dataset and their internals can be exploited for few-shot discriminative tasks. Yet, there is a loss of information in the process of learning the data distribution which we quantify based on the ability to project samples into the latent space of GANs and the loss in performance of constrastive models trained on the GAN generated images.
The next few sections describe these tradeoffs in detail.

\begin{table*}[]
\centering
\begin{tabular}{c|ccccc}

Method            & Face34     &   Face34 weighted    & Face8            & Car20            & Cat16            \\ \shline
Transfer Learning & 0.4577 $\pm$ 0.0151 & - & 0.6283           & 0.3391 $\pm$ 0.0094 & 0.2252 $\pm$ 0.0061 \\ 
Semi-Supervised~\cite{mittal2019semi}   & 0.4817 $\pm$ 0.0066 & - & 0.6336           & 0.4451 $\pm$ 0.0057 & 0.3015 $\pm$ 0.0035 \\ 
DatasetGAN~\cite{zhang2021datasetgan}     & 0.5346 $\pm$ 0.0121 & - & 0.7001 & 0.6233 $\pm$ 0.0055 & 0.3126 $\pm$ 0.0071 \\ 
DatasetGAN$^*$     & 0.5365 $\pm$ 0.0043 & 0.7959 $\pm$ 0.0237 & 0.6971 $\pm$ 0.0095 & 0.6840 $\pm$ 0.0322 & 0.3083 $\pm$ 0.0068 \\ 
\begin{tabular}[c]{@{}c@{}}Ours - CONV + distill\end{tabular} &
  \textbf{0.5406 $\pm$ 0.0015} &
  \textbf{0.8241 $\pm$ 0.0133}&
  \textbf{0.7048 $\pm$ 0.0028} &
  \textbf{0.6956 $\pm$ 0.0178} &
  \textbf{0.3101 $\pm$ 0.0042} \\ 
\end{tabular}
\caption{\textbf{Comparison of our best method with prior art:} We report mean IOU and standard deviation of each approach for various datasets. Face8 is an out of domain experiment where the training set is same as that of Face34 but the testing set is from a different dataset. Our approach with $f_{PROJ}$ as CONV with distillation outperforms previous methods in all of the test settings. DatasetGAN$^*$ denotes results after re-implementation using their updated code, where they change their upsampling technique and uncertainty calculation.}
\vspace{-3px}
\label{tab:usvsthem}
\end{table*}

% \usepackage{graphicx}
\begin{table*}[]
\centering
\resizebox{\textwidth}{!}{%
\begin{tabular}{cc|cc|cc|cc|cc|cc|
>{\columncolor[HTML]{D9E9FF}}c 
>{\columncolor[HTML]{D9E9FF}}c }
\multicolumn{2}{c|}{Method} &
  \multicolumn{2}{c|}{Face34} &
  \multicolumn{2}{c|}{Face34 weighted} &
  \multicolumn{2}{c|}{Face8} &
  \multicolumn{2}{c|}{Car20} &
  \multicolumn{2}{c|}{Cat16} &
  \multicolumn{2}{c}{\cellcolor[HTML]{D9E9FF}ReGAN/Face10} \\
\multicolumn{1}{c|}{Pretrain} &
  Projector &
  Direct &
  Distill &
  Direct &
  Distill &
  Direct &
  Distill &
  Direct &
  Distill &
  Direct &
  Distill &
  Direct &
  Distill \\ \shline
\multicolumn{1}{c|}{StyleGAN} &
  MLP &
  0.4301 &
  0.5365 &
  0.7572 &
  0.7959 &
  0.6331 &
  0.6971 &
  \multicolumn{1}{l}{0.5153} &
  0.6840 &
  0.1585 &
  0.3083 &
  0.7624 &
  0.7505 \\
\multicolumn{1}{c|}{MoCo} &
  MLP &
  0.5162 &
  0.5341 &
  0.7784 &
  0.7953 &
  0.6874 &
  0.7052 &
  0.6143 &
  0.6788 &
  0.2902 &
  0.3048 &
  0.7509 &
  0.7737 \\
\multicolumn{1}{c|}{MoCo} &
  CONV &
  0.5414 &
  0.5406 &
  0.8038 &
  0.8241 &
  0.6994 &
  0.6980 &
  0.6447 &
  0.6956 &
  0.3021 &
  0.3101 &
  0.7683 &
  0.7811
\end{tabular}%
}
\caption{\textbf{Detailed comparison of the effect of each step on mIoU :} The first row shows scores for generative models where "Direct" is obtained by performing latent optimization on test images to predict masks using obtained latent and "Distill" is the score after distillation. The other two rows are scores of contrastive models with different architectures of $f_{PROJ}$. The blue shaded column is the comparison with ReGAN method, where we use different architectures for the GAN method and different train/test settings following their paper.}
\label{tab:detailedcomparison}
\vspace{-12px}
\end{table*}

\section{\textsc{GanOrCon:} Generative or Contrastive}\label{sec:method}

We present a contrastive learning framework for few-shot fine-grained part segmentation. We construct our approach in phases similar to GAN-based segmentation methods as presented in DatasetGAN~\cite{zhang2021datasetgan} and ReGAN~\cite{tritrong2021repurposing}, so as to directly compare various aspects such as computational complexity, effectiveness, labeling difficulty and robustness. In the following paragraphs, we compare the construction of the GAN and CON methods for each phase. The steps include I) a representation learning framework (generative or contrastive), II.0) extracting hypercolumns from the trained model, II) training a few-shot projector over feature activation maps learnt in the previous step, and III) a supervised distillation training step using examples generated from the combined model of steps I and II.

\paragraph{Setup:} Let $\mathbf{x} \in \mathbb{R}^{H \times W \times 3}$ denote an image of an object and $\mathbf{y} \in \mathbb{M}^{H \times W}  |  \mathbb{M} = \{1, \dots, C\}$ denote the pixel-wise classification labels, with  $C$ denoting the number of classes.
Given a large number of unlabeled examples $\{\mathbf{x}\}$ and a limited number of labeled examples $\{(\mathbf{x},\mathbf{y})\}$, the goal is to learn a mapping function $\mathbf{f}(\mathbf{x}): \mathbf{x} \rightarrow \mathbf{y}$ that generates a pixel-wise classification label.

\vspace{-10px}
\paragraph{Step I -- Representation Learning:} 
Given a large set of unlabeled images $\{\mathbf{x} | \mathbf{x} \sim p_{data}(\mathbf{x})\}$, a generative model ($f_{GEN}$) is trained to map from a noise distribution $p_z(\mathbf{z})$ to the real data space via a min-max objective:
\begin{equation} \label{eq:min-max}
\vspace{-6px}
    \min_{f_{GEN}} \max_{f_D} \mathbb{E}_{\mathbf{x} \sim p_{data}(\mathbf{x})}[f_D(\mathbf{x})] - \mathbb{E}_{\mathbf{z} \sim p_{z}(\mathbf{z})}[f_D(f_{GEN}(\mathbf{z}))]
\end{equation}
where $f_{GEN}$ denotes the generator network and $f_D$ the discriminator network. The generator tries to mimic a real image while the discriminator tries to distinguish between a real image and an image from the generator network.

On the other hand, in our approach we train a contrastive learning model $f_{CON}$ which learns underlying representations by differentiating between images in the same unlabeled dataset. We experiment with methods that work both with (MoCo) and without (SimSiam) negative examples. 

The Momentum Contrast (MoCo~\cite{he2020momentum}) method minimizes the Noise-Contrastive Estimation (InfoNCE~\cite{oord2018representation}) loss.
Given a sample $\mathbf{x}$ and its transformation $\mathbf{q}_+$ as well as other samples $\mathbf{q}_i$, $i \in \{1,2...N\}$, such that $\mathbf{q}_+ \in  \mathbf{q}_i$, InfoNCE is defined as:
\begin{equation}
\label{eq:infonce}
\vspace{-6px}
\mathcal{L}_{\mathit{InfoNCE}} = -\log\frac{\exp\left(\langle f_{CON}(\mathbf{x}),f_{CON}(\mathbf{q}_+)\rangle\right)}{\sum_{i=1}^N\exp(\langle f_{CON}(\mathbf{x}), f_{CON}(\mathbf{q}_i)\rangle)}
\end{equation}
Here $\mathbf{q}_+$ is generated by applying photometric and geometric transformations on the anchor sample $\mathbf{x}$. We use MoCo V2~\cite{chen2020improved} in our experiments which uses the above loss.

Simple Siamese (SimSiam~\cite{chen2021exploring}) network trains without the use of negative examples, wherein given an additional predictor network ($h$), it minimizes the cosine feature distance between two views ($\mathbf{x}$, $\mathbf{q}_+$) of the same image using:
\begin{equation}
\label{eq:cosine}
\mathcal{L}_{\mathit{cosine}} = - \frac{h(f_{CON}(\mathbf{x}))}{\|h(f_{CON}(\mathbf{x}))\|_2} . \frac{f_{CON}(\mathbf{q}_+)}{\|f_{CON}(\mathbf{q}_+)\|_2}
\end{equation}

Networks trained using the above methods complete the first part of the method. Next we extract hypercolumns~\cite{hariharan2015hypercolumns} to learn segmentation maps from. 

\vspace{-10px}
\paragraph{Step II.0 -- Hypercolumn Extraction:} A deep network of $n$ layers (or blocks\footnote{Due to skip-connections, we cannot decompose the encoding over layers, but can across blocks.}) can be written as $\Phi(\mathbf{x}) = \Phi^{(n)} \circ \Phi^{(n-1)} \circ\dots\circ \Phi^{(1)}(\mathbf{x})$.
A representation $\Phi(\mathbf{x})$ of size $H' \times W' \times K$ can be spatially interpolated to input size $H\times W \times K$ to produce a pixel representation $\Phi_{i}(\mathbf{x}) \in \mathbb{R}^{H \times W \times K}$.
The hypercolumn representation of layers $l_1, l_2, \dots, l_n$ is obtained by concatenating interpolated features from corresponding layers i.e.\  $\Phi_{i}(\mathbf{x}) = \Phi^{(l_1)}_{i}(\mathbf{x}) \oplus \Phi^{(l_2)}_{i}(\mathbf{x})\oplus \dots \oplus \Phi^{(l_n)}_{i}(\mathbf{x})$. Both generative and contrastive methods extract hypercolumns from their respective trained models for few-shot segmentation. For StyleGAN this is comprised of the outputs of the AdaIN layers.

\vspace{-10px}
\paragraph{Step II -- Few-shot Segmentation:}
Using the extracted $\Phi_i(\mathbf{x})$, both approaches train a projector network ($f_{PROJ}$) to obtain pixel-wise confidence maps of shape $H\times W \times C$ where $C$ denotes the number of classes.

DatasetGAN utilizes an ensemble of MLP classifiers as $f_{PROJ}$ and use majority voting at test time\footnote{DatasetGAN also uses this ensemble to predict uncertainty for filtering generated data before the distillation step.}. Our method is more simpler in comparison. We use two variations for the architecture of $f_{PROJ}$, one being a MLP and the other a modified UNet~\cite{ronneberger2015u} (see Appendix \ref{sec:architectures} for architectures) which we will call CONV hereon.

At this point, we can evaluate on test images directly to obtain our performance measure, however to compare with the generative method we also perform distillation.

\vspace{-10px}
\paragraph{Step III -- Supervised Distillation:}
To infer on unseen images, generative models can use two approaches 1) latent code optimization to obtain latents which can nearly estimate the test image, 2) perform distillation using a standard supervised segmentation network. Since the first method affects performance (see Tab. \ref{tab:detailedcomparison}) and it is time consuming to obtain latents for test images, the second method is preferred by DatasetGAN. For generative method, the optimal solution of the distillation process can be formulated as:
\begin{equation} \label{eq:gandistill}
\vspace{-4px}
     \mathcal{\theta}^{*}_{\mathit{S}} = 
     \operatorname*{argmin}_{\theta_{\mathit{S}}}
     \ \ \mathbb{E}_{\mathbf{z}}[\mathbf{CE}(T(\mathbf{z};\theta_{T}), S(f_{GEN}(\mathbf{z});\theta_{S}))]
\end{equation}

where $\mathit{S}$ is the Student network with parameters $\mathit{\theta_{S}}$, $\mathit{T}$ is the teacher network with parameters $\mathit{\theta_{T}}$ and $\mathbf{CE}$ denotes the Cross Entropy loss. For the generative method, $\mathit{T(\mathbf{z};\theta_{T})}$ involves generating images from random latents $\mathbf{z}$ using $f_{GEN}$ and then obtaining corresponding masks using the Interpreter network.

In our case, $\mathit{T(\mathbf{x};\theta_{T})}$ is inferencing over our trained contrastive  + projector networks with images $\mathbf{x}$ from a pre-existing dataset to obtain masks. The student parameters can be optimized as below:
\begin{equation} \label{eq:condistill}
\vspace{-4px}
     \mathcal{\theta}^{*}_{\mathit{S}} = 
     \operatorname*{argmin}_{\theta_{\mathit{S}}}
     \ \ \mathbb{E}_{\mathbf{x}}[\mathbf{CE}(T(\mathbf{x};\theta_{T}), S(\mathbf{x};\theta_{S}))]
\end{equation}

It is important to note that the distillation step is optional for our method since it can directly infer on any given test image. However this step offers a little improvement in performance in some cases.

\subsection{Design considerations}\label{sec:design}
\vspace{-2px}
\subsubsection{Computational Complexity}
\label{sec:computations}
\vspace{-3px}
The computational bottleneck for the GAN based methods is the StyleGAN~\cite{karras2019style} training, whereas for our case its the contrastive learning network. According to the official repository of StyleGAN, it takes 11 days and 8 hours to train on 70000 images of resolution 1024$\times$1024 on 4 Tesla V100 GPUs. Our network trained on the same dataset takes 7 days and 10 hours on 4 1080TI GPUs. By crude linear scaling using the ratios of RAM and FLOPs of the GPUs, we can infer that our method is around ~7$\times$ faster.

We also compare with a StyleGAN2~\cite{karras2020analyzing} based method. In this case, our method is even more computationally efficient. StyleGAN2 takes 18 days and 14 hours on the same dataset and setting as discussed above. Thus our unsupervised training is ~11$\times$ faster in comparison.

The second step of both the methods i.e.\ the Style Interpreter training of DatasetGAN and Projector network training of our method have similar complexity as the input feature dimensions and network architectures are similar. Our method takes lesser time compared to DatasetGAN as they use an ensemble of classifiers. Note that this step takes the least amount of time as number of examples is very limited.

Furthermore, StyleGAN based methods need the distillation step to infer on unseen images. The StyleGAN + interpreter model first needs to generate a large number of images. The data generation and training of the distillation network takes $\sim$13 hours on 4 1080ti GPUs with DatasetGAN's settings for 10k training images. Our method can avoid this step and still produce competitive results.

\vspace{-10px}
\subsubsection{Effectiveness}
\vspace{-6px}
We evaluate our method on the same train and test settings as DatasetGAN for various datasets. We also compare with ReGAN for face data. Our method using MLP is able to achieve similar results w.r.t. the GAN based approaches after distillation, while our CONV based method outperforms the baselines both with and without (except Cars data) distillation. We discuss our results in greater detail in \S\ref{sec:results}.

\vspace{-10px}
\subsubsection{Labeling Difficulty}
\vspace{-6px}
GAN based approaches suffer from the following two major drawbacks when it comes to training and inference of few-shot segmentation, which our approach is able to avoid.
\vspace{-6px}
\begin{enumerate}
    \item To create training examples for the few-shot setting, generative model based methods need manual annotations on images generated by the GAN. This is due to the fact that latent optimization is not very accurate and is especially taxing for a pixel-wise classification task. On the other hand, our approach can use any existing dataset to train the segmentation projector network.
\vspace{-6px}    
    \item To evaluate GAN based segmentation network directly, we again need to perform latent code optimization. We show in Tab.~\ref{tab:detailedcomparison} that this creates a gap in the final performance. Thus, to get a fair evaluation of the GAN based methods an extra step of distillation is required. This requires the GAN to generate a large number of images and labels ($\sim$10k) to train a supervised segmentation network. 
 
 \vspace{-6px}

\end{enumerate}

\vspace{-12px}
\subsubsection{Robustness}
\begin{figure}
    \centering
    \includegraphics[width=0.9\linewidth]{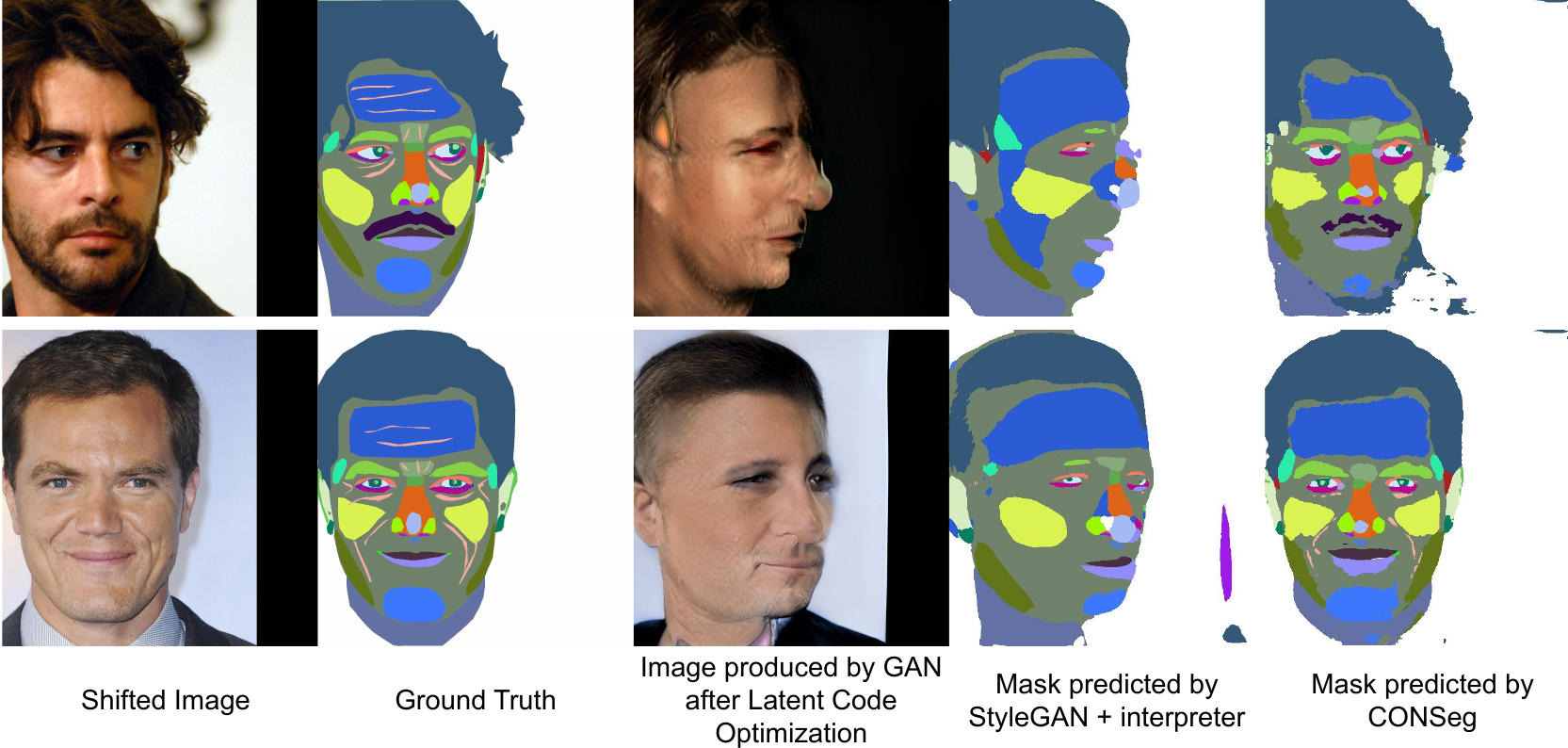}
    \caption{\textbf{Robustness of GAN vs CON :} Generative method is not able to model test images when some perturbation is introduced. The third column shows the resultant images after 10k iterations of latent code optimization. The contrastive model is able to deal with the shift in the images, wherein the resultant masks are shifted similarly to the input as visualized in the last column.}
    \label{fig:robust}
\end{figure}

\vspace{-6px}
Representations learned by generative modeling techniques are very sensitive to affine transforms. GANs will not be able to produce images which have a small perturbation w.r.t. their learnt representation space such as translation. Our approach is robust to such affine transforms and outputs a segmentation map with affine corresponding to the input image. We show this by shifting the test images and inferring our model on them. Fig.~\ref{fig:robust} shows our results on the transformed data. To get segmentation maps from the GAN based network for the translated images, we perform latent optimization on them. Fig. \ref{fig:robust} shows that the generative model is not able to map these shifted images. Since the test images are shifted to the left, the StyleGAN generates images where the head is turned to the right, which is a bias of the underlying distribution that it has learned from the training data. The GAN based methods thus rely on the extra distillation step to overcome this disadvantage.

\begin{table}[]
\centering
\begin{tabular}{c|cccc}
Iterations           & 1000                 & 5000                 & 10000                & 20000                \\ 
\shline
IoU                  & 0.4178               & 0.4270               & 0.4302               & 0.4283               \\ 
\end{tabular}
\vspace{-4pt}
\caption{\textbf{Direct evaluation of StyleGAN + interpreter :} mIoU vs iterations of latent optimization on images of Face34 test data.}
\label{tab:lco}
\vspace{-10px}
\end{table}

\vspace{-4pt}
\section{Experiments}
In this section we present the experimental details of our approach. We outline the datasets used, implementation details for each step, evaluation metrics and the settings for comparison to baseline methods.
\vspace{-4pt}
\subsection{Datasets}
\vspace{-3pt}
\subsubsection{Human Faces} 
\vspace{-8pt}

\textbf{Training:} 
For training the contrastive learning models, we use the Flicker-Faces-HQ~\cite{karras2019style} and the CelebAMask-HQ~\cite{liu2015deep} datasets. They have 70k and 30k images respectively, of 1024$\times$1024 resolution. For all training, we resize images to 512$\times$512 unlike StyleGAN which uses the original larger size. For the few shot segmentation training part, we use all 16 images from the Face34 training dataset released by DatasetGAN~\cite{zhang2021datasetgan}. The number in the names of the datasets correspond to the number of classes. We then choose a subset comprising of 10k images from CelebAMask-HQ to perform inference of our trained model. These images paired with the corresponding predicted masks form the training set for distillation. For all experiments of DatasetGAN we generate 10k images as well.

We also compare our method with ReGAN, where we train our models using the same datasets as their method. We use our unsupervised learning model that is trained on the Flicker-Faces-HQ dataset and train the few shot segmentation part using the same subset of 10 images from CelebAMask-HQ as ReGAN. Correspondingly, we convert the labels to 10 classes from 19 (see Appendix~\ref{sec:label}). We refer to this 10 class train/test data as Face10. For distillation we use the same subset of 10k images as earlier.

\noindent\textbf{Testing:} 
We use the testing set with 20 images of Face34 to compare to DatasetGAN's corresponding scores. DatasetGAN also reports scores on Face8 which is a modified version of CelebAMask-HQ with 8 classes. We use the same label transformation (see Appendix~\ref{sec:label}) and sample 50 images randomly from CelebAMask-HQ's testing set of 2k images to do the evaluation for our method and baselines. For comparing with ReGAN we use a subset of 190 images from the training set of CelebAMask-HQ following their method. For all evaluation experiments, we resize images to 512$\times$512 resolution like both the GAN methods.
\vspace{-10pt}
\subsubsection{Cats}
\vspace{-6pt}
\textbf{Training:} 
We use LSUN Cats dataset~\cite{yu2015lsun} for training our unsupervised network. Unlike StyleGAN which uses the whole dataset (70M images) to train, we use only a subset of 100k images and still achieve better results. We resize images to 256$\times$256 resolution for all training runs. For the few shot segmentation part we use Cat16 dataset released by DatasetGAN. We filter out few examples which have visible annotation errors and finally use 28 images for training.

\noindent\textbf{Testing:}
We use the Cat16 testing set which has 20 images and resize them to 256$\times$256 to calculate IoU.

\vspace{-10pt}
\subsubsection{Cars}
\vspace{-6pt}
\textbf{Training:} 
Similar to the setting of Cats dataset, we use a subset of 100k images from 46M total images in LSUN Cars dataset~\cite{yu2015lsun} to train our contrastive model.  We use DatasetGAN's Car20 training set comprising of 33 images for training the few shot model. In this case, we resize to 512$\times$512 for both unsupervised and few shot training runs.

\noindent\textbf{Testing:}
We use 10 images from the Car20 test set and resize to 512$\times$512 similar to DatasetGAN for evaluation.

\subsection{Implementation Details}
\vspace{-6pt}
\paragraph{Contrastive Learning.} For unsupervised contrastive learning, we use two different frameworks namely MoCoV2~\cite{chen2020improved} and SimSiam~\cite{chen2021exploring}. We use ResNet50~\cite{he2016deep} as the backbone architecture for both of these approaches. We train MoCoV2 for 800 epochs and SimSiam for 400 epochs. We use a batch size of 32 for the training where input images are resized to 512$\times$512 (faces and cars) and 128 for the 256$\times$256 setting (cats).

\begin{table}[]
\centering
\begin{tabular}{c|cc}
Method         & MoCo   & SimSiam \\ 
\shline
MLP            & 0.5162 & 0.5038  \\ 
CONV           & 0.5414 & 0.5297   \\ 
MLP + distill  & 0.5341 & 0.5188       \\ 
CONV + distill & 0.5406 & 0.5173
\end{tabular}
\caption{\textbf{Effect of contrastive model on downstream IoU:} We report mIoU for the Face34 dataset for each method varied over the architecture of $f_{PROJ}$ and before and after distillation.}
\label{tab:mocovssimsiam}
\end{table}

\begin{table}[]
\centering
\begin{tabular}{c|cccc}
\multirow{3}{*}{Projector} & \multicolumn{4}{c}{Pretraining Strategy}                                                         \\ 
 &
  \multicolumn{1}{c|}{\begin{tabular}[c]{@{}c@{}}ImageNet\\ classification\end{tabular}} &
  \multicolumn{1}{c}{\begin{tabular}[c]{@{}c@{}}Imagenet\\ MoCo\end{tabular}} &
  \multicolumn{1}{c}{\begin{tabular}[c]{@{}c@{}}FFHQ\\ MoCo\end{tabular}} &
  \begin{tabular}[c]{@{}c@{}}CelebA\\ MoCo\end{tabular} \\ \shline
MLP                        & \multicolumn{1}{c|}{0.1893} & \multicolumn{1}{c}{0.4650} & \multicolumn{1}{c}{0.5023} & 0.5162 \\
CONV                       & \multicolumn{1}{c|}{0.2551} & \multicolumn{1}{c}{0.4828} & \multicolumn{1}{c}{0.5132} & 0.5414
\end{tabular}
\caption{\textbf{Effect of pretraining of $f_{CON}$ on mIoU :} The first column reports scores using ResNet50 trained on ImageNet1K classification task. The other columns compare MoCo trained with different datasets with the same ResNet50 backbone. All scores are on the Face34 test set without distillation.}
\label{tab:pretraining}
\vspace{-10pt}
\end{table}

\vspace{-14pt}
\paragraph{Few Shot Segmentation.} After obtaining the trained ResNet50 we freeze its weights and extract hypercolumns by stacking activations of the layers from \texttt{conv\textunderscore2x} to \texttt{conv\textunderscore5x}. We upsample all the hypercolumns to the same resolution as the input image and concatenate them. We use the resultant 3D feature tensor to train our projector network. We use two different approaches : MLP and CONV as our $f_{PROJ}$. We use random resized crop, flip and color jitter as augmentations. We train the MLP for 800 epochs with an initial lr of 0.001 and cosine lr decay. We train the UNet based architecture - CONV, for 200 epochs with an initial lr of 0.0005. During this training, in the ResNet50 architecture we change the stride of the first convolution layer from 2 to 1 while extracting hypercolumns to train the projector model in order to preserve more spatial information.

\vspace{-12pt}
\paragraph{Supervised Distillation.} Once the few shot segmentor is trained, we use it to create a dataset to perform distillation. We use Deeplabv3~\cite{chen2017rethinking} with ResNet101 as the student model for all experiments. We use an initial lr of 0.001 with a batch size of 8 and train for 2 epochs for every setting.

\begin{figure*}
    \centering
    \includegraphics[width=0.93\linewidth]{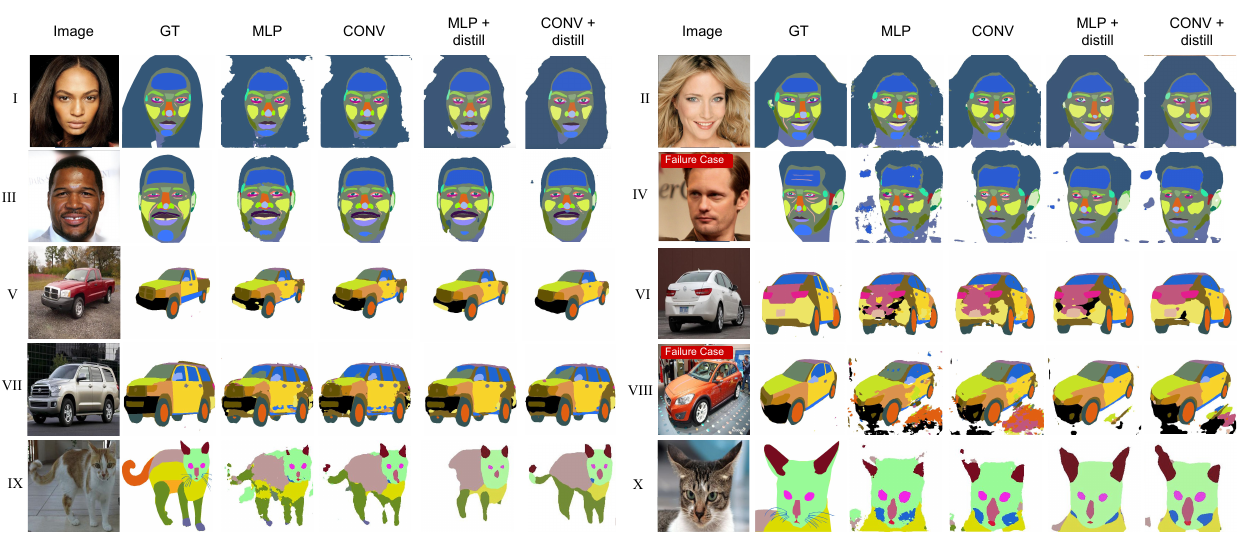}
    \caption{\textbf{Qualitative Results on various datasets:} We compare the visual results for our MLP and CONV methods both with and without distillation. We present results on the test sets of Face34, Car20 and Cat16 datasets. In image IV and VIII, we show two failure cases.}
    \label{fig:viz}
    \vspace{-8pt}
\end{figure*}

\vspace{-4pt}
\subsection{Evaluation}
\vspace{-6pt}
We follow the same evaluation strategy as DatasetGAN for all scores that we report. The testing set is split into five folds and each set is chosen as validation set to choose the best checkpoint for testing. We finally report the mean Intersection over Union (mIoU) over all the folds.

We also carry out a comparison of weighted mIoU for human face datasets, consistent with ReGAN's evaluation method. We compute the weight of each class as the ratio of the number of ground-truth pixels belonging to the class to the total number of pixels. We then perform the same cross validation as described above.

\vspace{-4pt}
\subsection{Comparison to Baselines}
\vspace{-5pt}
\paragraph{Direct Evaluation of StyleGAN segmentor.}
DatasetGAN's method has three training steps : StyleGAN, Style interpreter and parts segmentation network (DeepLabV3). We want to evaluate their method at the second step i.e.\ infer using the StyleGAN + interpreter network with our test set to produce segmentation masks and then compute mIoU. To do so we first need to find the latents which most nearly produce the test images. We perform latent code optimization with different number of iterations to project test images into the latent space. We then pass these resultant latents to the combined network and obtain predictions to compute the mean IoU. Tab. \ref{tab:lco} shows the variation of mIoU with number of iterations of latent code optimization. It can be seen that there is a considerable gap in these scores and the one after the supervised distillation step (Tab. \ref{tab:usvsthem} \& \ref{tab:detailedcomparison}).

\vspace{-13pt}
\paragraph{Face8 and Face10.}
We compare with DatasetGAN on Face8 which is a modified version of CelebAMask-HQ dataset with 8 classes instead of 19. For these experiments we do not retrain our model on CelebA. We use the existing model trained on Face34 and merge labels after inference to form the 8 classes from 34, for both GAN and CON. 
% We do the same to produce scores for DatasetGAN also.

The Face10 values are in comparison with the ReGAN~\cite{tritrong2021repurposing} paper. In this case, we need to retrain our model on CelebA examples modified to 10 classes since one of the classes ``cloth" is not a part of the classes in Face34. We use the same 10 train images as used by ReGAN to compare to their 10-shot setting. We keep our network architectures same, whereas to replicate ReGAN's numbers we use the architectures and hyperparameters as specified in their paper. We use weighted IoU for evaluation following their paper. Tab.~\ref{tab:detailedcomparison} shows that ReGAN performs better in the direct step as compared to after distillation. This may be attributed to the fact that StyleGAN2 works better with latent code optimization. Both our methods are able to outperform ReGAN after distillation. Our CONV variant performs better in the direct phase too.

\begin{table}[]
\centering
\begin{tabular}{c|cc}

Training set         & Cars   & Faces  \\ \shline
GAN generated images & 0.4266 & 0.3868     \\ 
Real images          & 0.6143 & 0.5162
\end{tabular}
\caption{mIoU scores on Face34 and Car20 for MoCo trained with images generated from StyleGAN vs from real dataset.}
\label{tab:generatedvsreal}
\vspace{-10pt}
\end{table}

\vspace{-7pt}
\section{Results and Discussion}
\label{sec:results}
\vspace{-5pt}
In this section we present comparison of our quantitative performance with prior art, discuss on latent code optimization, perform ablation over various training settings and analyze the qualitative results of our various methods.
\vspace{-4pt}
\subsection{Performance comparison with other methods}
\vspace{-4pt}
 Tab.~\ref{tab:usvsthem} compares the performance of our best method after distillation with transfer learning, semi-supervised learning~\cite{mittal2019semi} and DatasetGAN baselines. The transfer network was pre-trained on the semantic segmentation task of MS-COCO~\cite{lin2014microsoft}. Our method performs better than the baseline methods on all the datasets.

 Tab.~\ref{tab:detailedcomparison} presents a detailed comparison of our method with the GAN methods. For Face34 we also compare with weighted mIoU, the metric used in ReGAN. Distillation improves performance in all cases, the highest jump can be seen in the Car20 dataset. Except ReGAN, in all cases the direct evaluation of GAN segmentor yields much worse results than after distillation. 
 
 We compare with ReGAN in the shaded column with weighted mIoU. In this case the direct method performs better than the distilled. This is due to the fact that StyleGAN2 is better able to work with latent code optimization, however the issues of robustness and inferring on unseen images will still remain. StyleGAN2 is also even more computationally expensive as compared to StyleGAN (\S\ref{sec:computations}). 
 
 Our variant with CONV as the chosen $f_{PROJ}$ architecture outperforms the MLP method, except for the Face8 setting. This method achieves better mIoU scores than the GAN baseline both with and without distillation.

\vspace{-2pt}
\subsection{Difficulty of latent code optimization}
\vspace{-4pt}
We analyze the performance of direct evaluation of DatasetGAN's StyleGAN + Style Interpreter network on unseen images from the Face34 test set. We vary the number of iterations of latent code optimization to find corresponding latents for the StyleGAN network using the input test images. In Tab.~\ref{tab:lco} we show the variation of mIoU with iterations. The score does not increase after 0.4302 at 10k iterations. There is a considerable gap between this number and the score of 0.5365 mIoU obtained by DatasetGAN after distillation. This is due to the fact that latent code optimization is not able to produce images as close to the input image as is required for reasonable performance in fine-grained segmentation.

\vspace{-2pt}
\subsection{Training of Backbone network}
\vspace{-3pt}
In this section we ablate on various training stategies for our backbone network $f_{CON}$. For all the following experiments, we keep the architecture fixed to that of ResNet50.

\vspace{-12pt}
\paragraph{MoCo vs SimSiam.}
We compare contrastive learning frameworks which train with (MoCoV2) and without (SimSiam) negative examples. Tab.~\ref{tab:mocovssimsiam} compares the scores of each with the MLP model as $f_{PROJ}$ both before and after distillation. MoCoV2 is able to achieve better performance on the few-shot segmentation task than SimSiam.  However, SimSiam needs half as many epochs (400 epochs) to reach the reported score compared to MoCoV2 (800 epochs).

\vspace{-12pt}
\paragraph{Comparison with Imagenet1K pretrained.}
The first column in Tab.~\ref{tab:pretraining} shows the mIoU on Face34 dataset for backbone network $f_{CON}$ initialized using Imagenet1K~\cite{krizhevsky2012imagenet} classification weights. The performance boost that MoCo pre-training offers is significant.

\vspace{-12pt}
\paragraph{MoCo with different real datasets.}
The rest of the columns of Tab.~\ref{tab:pretraining} compares the effect of the unlabeled dataset. We obtain the best performance using CelebA since the test images belong to CelebA's test set. FFHQ allows better performance than ImageNet because it is comprised solely of faces. However it is interesting to note that even though ImageNet dataset has no face images, the segmentor using MoCo trained on it still obtains good scores, comparable to FFHQ/CelebA.

\vspace{-12pt}
\paragraph{MoCo training on GAN generated images.}
To test the bias of generative models, we train MoCo on images generated by StyleGAN. We use the same resolution and number of unlabeled images for both training runs. Tab. \ref{tab:generatedvsreal} lists the scores on Face34 and Car20. There is a considerable drop in mIoU for the MoCo trained on images generated by StyleGAN. The MoCo training on generated images also reaches high accuracy much earlier ($\sim$ 90$\%$ at 200 epochs). This indicates that GANs are not able to model a large domain i.e\ they are not able to learn a general distribution. This might also mean that GAN generated images have some typical artefacts that CON models are able to identify.

\vspace{-2.5pt}
\subsection{Qualitative Results}
\vspace{-3pt}

Fig. \ref{fig:viz} presents a qualitative comparison of the strengths and weaknesses of our methods. In image I, because of the black background both the direct MLP and CONV models struggle with segmenting the hair, though the CONV does a better job. In this case distillation helps improve the predicted mask. For images II and III, the CONV + distill variants are able to capture the overall structure better, whereas the MLP + distill methods are able to differentiate smaller regions such as wrinkles and chin. In image IV, the MLP model confuses a large percentage of the background as neck/forehead. The CONV method is able to mitigate this problem as it can take more global cues owing to its larger receptive field (see Appendix~\ref{sec:architectures}). In images V and VII, the CONV + distill approach is able to eliminate confusion between back and front bumpers which is present in the predictions of the other methods. In image VI, the MLP model makes some wrong predictions. The CONV model is able to clear up most of the confusion, while the CONV + distill model boosts this performance and minimizes faulty predictions. In image VIII, we show a failure case. Predictions on the cats data has suffered due to faulty annotations in the dataset and the large variation of pose and occlusion making it challenging to learn representations.

\vspace{-6px}
\section{Conclusion}
\vspace{-5px}
We present a contrastive learning approach for few-shot fine-grained part segmentation. We compare to previous generative approaches and show using various experiments that our method achieves better performance, is simpler, an order-of-magnitude faster and is more robust. GAN generated datasets while effective on the part-segmentation task, might not be as effective on tasks such as fine-grained image classification due to biases they introduce. Further investigation of the nature of these biases could better guide the community adoption of GANs as a replacement of datasets. 

\vspace{-10px}
\paragraph{Acknowledgements} The project is supported in part by Grant
\#1749833 from the National Science Foundation of United
States. Our experiments were performed on the University
of Massachusetts Amherst GPU cluster obtained under the
Collaborative Fund managed by the Mass. Technology Collaborative.
%%%%%%%%% REFERENCES
{\small
\bibliographystyle{ieee_fullname}
\bibliography{egbib}
}

\clearpage
\appendix

\onecolumn

\noindent{\Large \textbf{Appendix}}

\section{Dataset Labels}
\label{sec:label}
For experiments on the CelebA-HQ dataset, we transform labels manually from 19 classes to 8 or 10 as required for comparison to the baselines. The 19 classes in the original dataset are:

\begin{center}
\vspace{-.3em}
\tablestyle{1pt}{1.2}	
\begin{tabular}{c|ccccccccccccccccccc}
Index & 0 & 1 & 2 & 3 & 4 &5&6&7&8&9&10&11&12&13&14&15&16&17&18 \\
\shline
Class & `background' & `skin' & `nose'& `eye\_g'	& `l\_eye' & `r\_eye' & `l\_brow' & `r\_brow' & `l\_ear' & `r\_ear' & `mouth'	& `u\_lip' & `l\_lip' & `hair' & `hat' & `ear\_r' & `neck\_l' & `neck' & `cloth'
\end{tabular}
\vspace{-.5em}
\end{center}

\noindent For experiments with 10 classes and 8 classes we consider the following:

\begin{flushleft}
\vspace{-.1em}
\tablestyle{1pt}{1.2}	
\begin{tabular}{c|cccccccccc}
Index & 0 & 1 & 2 & 3 & 4 &5&6&7&8&9 \\
\shline
Class & `background' & `skin' & `nose'& `eye' & `brow' & `ear' & `mouth' & `hair' & `neck' & `cloth'
\end{tabular}
% \vspace{-.5em}
\quad
\vspace{-.1em}
\tablestyle{1pt}{1.2}	
\begin{tabular}{c|cccccccc}
Index & 0 & 1 & 2 & 3 & 4 &5&6&7 \\
\shline
Class & `background' & `skin' & `nose'& `eye' & `brow' & `ear' & `mouth' & `hair' 
\end{tabular}

\end{flushleft}

\noindent Note that in the case of transformation from 19 classes to 10, we assign classes 14,15,16 as the 'background'. For the case of transformation from 10 classes to 8, we assign classes 8,9 as the 'background'

\section{Network Architectures}
\label{sec:architectures}
For our MLP model we use a 2 layer fully connected network with hidden dimensions 1024 followed by 256. In the case of our modified UNet based model, we first downsample the hypercolumns and then upsample them to the input resolution. For faces and cars, which we process at 512 $\times$ 512 we use Model-A and for cats which we process at 256 $\times$ 256 we use Model-B. The architectures are as given below:

\vspace{1pt}
\begin{table}[h]
\begin{minipage}[c]{0.5\textwidth}
\centering

% \vspace{-2em}
\tablestyle{1.5pt}{1.2}
\caption*{Model-A}
\begin{tabular}{c|c|c}

Name   & Input Resolution & Layer                                                            \\
\shline
Conv1  & H$\times$W$\times$3840         & ConvBNReLU                                                       \\
MP1    & H$\times$W$\times$1024         & MaxPool                                                          \\
Conv2  & H/2$\times$W/2$\times$1024     & ConvBNReLU                                                       \\
MP2    & H/2$\times$W/2$\times$256      & MaxPool                                                          \\
Conv3  & H/4$\times$W/4$\times$256      & ConvBNReLU                                                       \\
MP3    & H/4$\times$W/4$\times$256      & MaxPool                                                          \\
Conv4  & H/8$\times$W/8$\times$256      & ConvBNReLU                                                       \\
MP4    & H/8$\times$W/8$\times$256      & MaxPool                                                          \\
Conv5  & H/16$\times$W/16$\times$256    & ConvBNReLU                                                       \\
MP5    & H/16$\times$W/16$\times$512    & MaxPool                                                          \\
Conv6  & H/32$\times$W/32$\times$512    & ConvBNReLU                                                       \\
Up1    & H/32$\times$W/32$\times$512    & \begin{tabular}[c]{@{}c@{}}Upsample\\ Concat(Conv5)\end{tabular} \\
Conv7  & H/16$\times$W/16$\times$1024   & ConvBNReLU                                                       \\
Up2    & H/16$\times$W/16$\times$256    & \begin{tabular}[c]{@{}c@{}}Upsample\\ Concat(Conv4)\end{tabular} \\
Conv8  & H/8$\times$W/8$\times$512      & ConvBNReLU                                                       \\
Up3    & H/8$\times$W/8$\times$256      & \begin{tabular}[c]{@{}c@{}}Upsample\\ Concat(Conv3)\end{tabular} \\
Conv9  & H/4$\times$W/4$\times$512      & ConvBNReLU                                                       \\
Up4    & H/4$\times$W/4$\times$128      & \begin{tabular}[c]{@{}c@{}}Upsample\\ Concat(Conv2)\end{tabular} \\
Conv10 & H/2$\times$W/2$\times$384      & ConvBNReLU                                                       \\
Up5    & H/2$\times$W/2$\times$256      & \begin{tabular}[c]{@{}c@{}}Upsample\\ Concat(Conv1)\end{tabular} \\
Conv11 & H$\times$W$\times$1280         & ConvBNReLU                                                       \\
FC     & H$\times$W$\times$256          & Linear                                                           \\
       & H$\times$W$\times$Classes      & -                                               
       
% \vspace{-3pt}
\end{tabular}

\end{minipage}
\begin{minipage}[c]{0.5\textwidth}
\centering
% \vspace{-0.1em}
\tablestyle{1.5pt}{1.2}
\caption*{Model-B}
\begin{tabular}{c|c|c}
Name  & Input Resolution & Layer                                                            \\
\shline
Conv1 & H$\times$W$\times$3840         & ConvBNReLU                                                       \\
MP1   & H$\times$W$\times$1024         & MaxPool                                                          \\
Conv2 & H/2$\times$W/2$\times$1024     & ConvBNReLU                                                       \\
MP2   & H/2$\times$W/2$\times$256      & MaxPool                                                          \\
Conv3 & H/4$\times$W/4$\times$256      & ConvBNReLU                                                       \\
MP3   & H/4$\times$W/4$\times$256      & MaxPool                                                          \\
Conv4 & H/8$\times$W/8$\times$256      & ConvBNReLU                                                       \\
MP4   & H/8$\times$W/8$\times$256      & MaxPool                                                          \\
Conv5 & H/16$\times$W/16$\times$256    & ConvBNReLU                                                       \\
Up1 & H/16$\times$W/16$\times$512 & \begin{tabular}[c]{@{}c@{}}Upsample\\ Concat(Conv4)\end{tabular} \\
Conv6 & H/8$\times$W/8$\times$768      & ConvBNReLU                                                       \\
Up2   & H/8$\times$W/8$\times$256      & \begin{tabular}[c]{@{}c@{}}Upsample\\ Concat(Conv3)\end{tabular} \\
Conv7 & H/4$\times$W/4$\times$512      & ConvBNReLU                                                       \\
Up3   & H/4$\times$W/4$\times$128      & \begin{tabular}[c]{@{}c@{}}Upsample\\ Concat(Conv2)\end{tabular} \\
Conv8 & H/2$\times$W/2$\times$384      & ConvBNReLU                                                       \\
Up4   & H/2$\times$W/2$\times$256      & \begin{tabular}[c]{@{}c@{}}Upsample\\ Concat(Conv1)\end{tabular} \\
Conv9 & H$\times$W$\times$1280         & ConvBNReLU                                                       \\
FC    & H$\times$W$\times$256          & Linear                                                           \\
      & H$\times$W$\times$Classes      & -                                                               
\end{tabular}

\end{minipage}
\end{table}

\newpage
\section{Re-implementation details}
\label{sec:reimplementation}
\textbf{DatasetGAN :} We present scores both reported by DatasetGAN in their paper and after using their updated code, where the upsampling technique for hypercolumns has been changed from 'nearest neighbour' to 'bilinear'. The mIoU of the Car category sees a large hike due to this. For the Cat category we filter out 2 images from the Cat16 training set with largely faulty annotations. We train DatasetGAN's model on the remaining images and report scores.

\textbf{ReGAN :} Since ReGAN have not released their code, we re-implemented their method as described in the paper. We use StyleGAN2 trained on FFHQ dataset. We implement their MLP model with 2 hidden layers of 2000 and 200 dimensions each. For distillation we use UNet architecture as specified in their paper. We compare with their 10-shot setting. We use the same 10/190 train/test split as communicated by the authors for both their method and ours.

\section{Hyperparameters}
\label{sec:hyperparameters}
\noindent 
In this section we list additional training details which we did not specify in the main text.

\textbf{MoCoV2 Training :}
We train MoCoV2 with MLP head with an initial lr of 0.03 and weight decay of 0.0001 for 800 epochs with cosine annealing. We use SGD with momentum of 0.9 and temperature coefficient for the loss as 0.2. For the transforms we use random resized crop, flip, gaussian blur, grayscale and color jitter.

\textbf{SimSiam Training :} 
We use the official implementation\footnote{\url{https://github.com/facebookresearch/simsiam}} of SimSiam. We train SimSiam with an initial lr of 0.05 and weight decay of 0.0001 for 400 epochs. We notice that training for more epochs (\eg 800) leads to slightly worse performance in the downstream tasks. We use SGD with momentum of 0.9 an same transforms as MoCoV2.

\textbf{Hypercolumn extraction:}
We use outputs of ResNet50 blocks from \texttt{conv\textunderscore2x} to \texttt{conv\textunderscore5x}. We use bilinear upsampling to resize to the resolution of input image before concatenation. The number of channels of our concatenated tensor is 3840.

\textbf{Projector Training :}
 We train for 800 epochs with initial learning rate of 0.001 for the MLP model, while for the UNet based methods we train for 200 epochs with initial lr of 0.0005. For both methods we use weight decay of 0.0005, Adam optimizer and batch size of 2. We also use random resized crop, flip and color jitter for both the methods.

\textbf{Distillation Training :} We use DeepLabV3 with ResNet101 for all experiments. We use initial lr of 0.001 and batch size of 8 with Adam optimizer. We observe that our models converge by the 2nd epoch and we use that model to report mIoUs. 

\section{Effect of Resolution}

\begin{table}[h]
\centering
\begin{tabular}{c|ccc}
\multicolumn{1}{l|}{} & \multicolumn{3}{c}{\begin{tabular}[c]{@{}c@{}}Input Resolution for\\  training MoCoV2\end{tabular}} \\
\begin{tabular}[c]{@{}c@{}}Input Resolution for \\ training segmentor\end{tabular} & 96 & 256 & 512 \\
\shline
96 & 0.4047 & 0.4081 & 0.3984 \\
256 & 0.4687 & 0.5022 & 0.4876 \\
512 & 0.4389 & 0.4958 & 0.5023
\end{tabular}
\end{table}

We explore the effect of resolution for each training on performance. Here we refer to the trained MoCoV2 + Hypercolumn extraction + Projector as the segmentor. While training the segmentor with input image of resolution $R_i$ we upsample the hypercolumns to and calculate loss on the same resolution $R_i$. While testing, we upsample the output of network to 512 $\times$ 512 for all experiments to obtain consistent scores. All MoCoV2 models have been trained on the FFHQ dataset, while the segmentor models have been trained on Face34. It can be seen that in this case there is almost no improvement in performance for MoCoV2 trained on 512 $\times$ 512/segmentor trained on 512 $\times$ 512, wrt MoCoV2 trained on 256 $\times$ 256/segmentor trained on 256 $\times$ 256. The performance of MoCo models trained on 96 $\times$ 96 is consistently worse.

\newpage
\section{Some more Qualitative Results}
We present more qualitative results of our best method on the Face34 and Car20 testing datasets below. In the case of faces, our model is able to capture all classes well except very fine segments such as some wrinkles. In the case of cars, our model classifies the window of the last example correctly, though it is wrongly annotated.
\vspace{-10pt}
\begin{figure*}[h]
    \centering
    \includegraphics[width=1.05\linewidth]{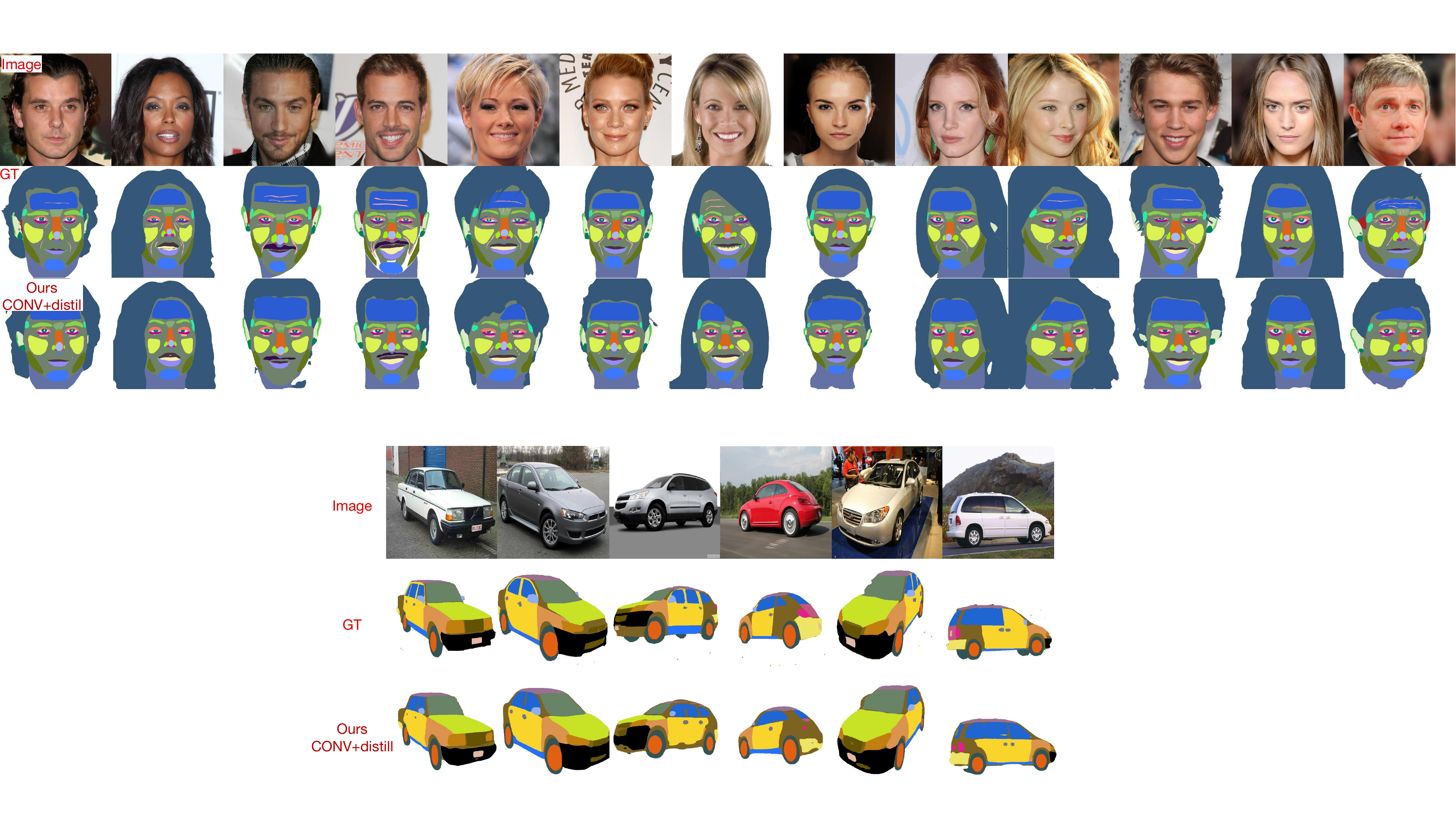}
    \label{fig:vizmore}
\end{figure*}

\end{document}